%% file: root.tex
\documentclass[letterpaper, 10 pt, conference]{ieeeconf}  % Comment this line out if you need a4paper

\IEEEoverridecommandlockouts                              % This command is only needed if 
\title{\LARGE \bf
Design and Control of a Micro Overactuated Aerial Robot\\ with an Origami Delta Manipulator
}

\author{Eugenio Cuniato$^{1}$, Christian Geckeler$^{2}$, Maximilian Brunner$^{1}$, Dario Str\"{u}bin$^{1}$, Elia B\"{a}hler$^{1}$, Fabian Ospelt$^{1}$,\\ Marco Tognon$^{1}$, Stefano Mintchev$^{2}$, Roland Siegwart$^{1}$% <-this % stops a space
\thanks{The research leading to these results has been supported by the AERO-TRAIN project, European Union's Horizon 2020 research and innovation programme under the Marie Skłodowska-Curie grant agreement No 953454, by the Swiss National Science Foundation through the Eccellenza Grant PCEFP2\_186865, by NCCR Digital Fabrication, Armasuisse Science and Technology and NCCR Robotics, a National Centre of Competence in Research, funded by the Swiss National Science Foundation (grant number 51NF40\_185543). The authors are solely responsible for its content.}% <-this % stops a space
\thanks{${}^{1}$Autonomous Systems Laboratory,
        ETH Zurich, Switzerland.}%
\thanks{${}^{2}$Environmental Robotics Laboratory,
ETH Zurich, Switzerland.}%
\thanks{Corresponding author email
       {\tt\small ecuniato@ethz.ch}}%
}

\input{symbol_definitions.tex}
\begin{document}

\maketitle
\thispagestyle{empty}
\pagestyle{empty}

%%%%%%%%%%%%%%%%%%%%%%%%%%%%%%%%%%%%%%%%%%%%%%%%%%%%%%%%%%%%%%%%%%%%%%%%%%%%%%%%
\begin{abstract}
    This work presents the mechanical design and control of a novel small-size and lightweight \ac{MAV} for aerial manipulation.
    %overactuated \ac{MAV} for aerial manipulation. 
    To our knowledge, with a total take-off mass of only \SI{2.0}{\kg},
    %, 25\% less than comparable state-of-the-art platforms, 
    the proposed system is the most lightweight \ac{AM} that has 8-DOF independently controllable: 5 for the aerial platform and 3 for the articulated arm.
    %This is obtained with a wise design of the system.
    %The aerial platform is designed to meet a good balance between actuation capabilities and weight. 
    \EC{We designed the robot to be fully-actuated in the body forward direction. This allows independent pitching and instantaneous force generation, improving the platform's performance during physical interaction.
        %The fully actuation is intentionally chosen in the body forward direction to allow independent pitching and instantaneous force generation, improving the platform's performance in physical interaction.
    }
    %
    % \EC{
    %     Thanks to its low weight and powerful actuation, the platform has a thrust-to-weight ratio of 2.43.
    % }
    %The platform is equipped with an origami delta manipulator driven by three servos, enabling active motion compensation at its end-effector.
    The robotic arm is an origami delta manipulator driven by three servomotors, enabling active motion compensation at the end-effector. Its composite multimaterial links help reduce the weight, while their flexibility allow for compliant aerial interaction with the environment. \EC{In particular}, the arm's stiffness can be changed according to its configuration.
    %The system has been specifically designed to fit inside confined spaces as small as 22 inches tubes. Strict initial requirements on flight time, payload and size are set. %Despite its small size and weight, the system is shown to be overactuated and suitable for contact-based inspection.
    %The vehicle is also equipped with a micro origami delta manipulator.
    We provide an in depth discussion of the system design and characterize the stiffness of the delta arm.
    %, the multi-layer manufacturing and the overall design are discussed. 
    A control architecture to deal with the platform's overactuation while exploiting the delta arm is presented.
    %This consists of a geometric pose controller for the \ac{MAV} and of an \ac{IK} controller specifically developed for the origami kinematic chain.
    Its capabilities are experimentally illustrated both in free flight and physical interaction,
    \EC{
        highlighting advantages and disadvantages of the origami's folding mechanism.
    }
\end{abstract}

%%%%%%%%%%%%%%%%%%%%%%%%%%%%%%%%%%%%%%%%%%%%%%%%%%%%%%%%%%%%%%%%%%%%%%%%%%%%%%%%
\section{INTRODUCTION}
\begin{figure}[t]
    \centering
    \includegraphics[width=1\columnwidth]{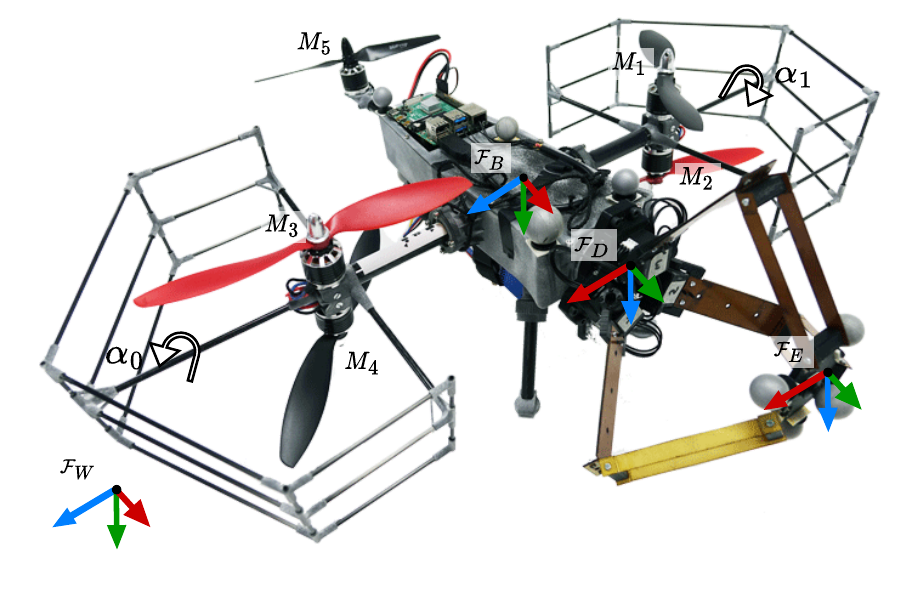}
    \caption{The overactuated \ac{MAV} with the origami delta manipulator. The body $\frameB$, world $\frameW$, delta base $\frameD$ and end-effector $\frameE$ frames are shown. The $\x$, $\y$, $\z$ axes are represented in red, blue, green, respectively. The arm tilt angles around the body $\B\y$ axis are indicated with $\alpha_0$, $\alpha_1$.}\label{fig:voliroPlusOrigami}
    \vspace{-1em}
\end{figure}
Nowadays, the interest for aerial platforms able to perform manipulation tasks is constantly increasing~\cite{tognonTRO}.
%
%with surfaces to analyze them, and determine if they are, for example, damaged or corroded.
%
Many inspection applications require specifically trained operators working at elevated locations performing interaction and manipulation tasks. %, that manually perform the desired measurements in the location of interest.
On the other hand, the use of \acp{AM} would reduce costs and operation time, improving safety as well. %identifying problems faster and even safer. % since the operators do not have to work in dangerous or remote places anymore.
%these controls are otherwise performed manually, which could expose the person performing the test to the risk of a fall if the structure to analyze is for example a bridge or a water tank.
%
The use of multi-directional thrust platforms has been proved fundamental to perform \ac{APhI} tasks, being able to independently exert forces and torques.
Different configurations of multi-directional thrust systems~\cite{hamandi2021design} have already been tested in the past: examples are tricopters~\cite{Papachristos2014}, quadcopters~\cite{Ryll2012,Junaid2018DesignAI}, hexacopters~\cite{kamel2018} and even octocopters~\cite{Brescianini2016}.
%
% Important systems parameters to consider for real applications are payload, flight time and efficiency~\cite{Bodie2018TowardsEF}.
%
%In addition, the change of power consumption with the platform orientation must also be taken into account~\cite{Allenspach2020}.

By enhancing the mobility of a \ac{MAV} with the dexterity of a robot manipulator, new possibilities are unlocked~\cite{2019-TogTelGasSabBicMalLanSanCorFra}.
%it can interact with the environment or perform inspection tasks whilst reaching confined spaces, previously inaccessible to conventional aerial platforms.
%
Among all possible robotic arms, delta manipulators are of particular interest for aerial manipulation because most of their weight is at the base, reducing the inertia and thus the reaction forces on the aerial platform during the motion. %dynamics. % then conventional manipulators.
Additionally, they possess very fast dynamics in their three translational \ac{DOF}, allowing them to compensate possible base position offsets or oscillations.
\EC{
    This made them a popular choice, both for \ac{APhI} with quadcopters~\cite{Fumagalli2014,chermprayong2019integrated,tzoumanikas2020aerial}, or end-effector tracking with an omnidirectional platform~\cite{Bodie2021a}.
}
However, due to the amount of required actuators, joints and linkages, the addition of actively driven end-effectors often results in a large and heavy system.
The work in~\cite{Meng2020} gives an overview of several \ac{AM} designs.
\EC{
    In particular, it shows that $60$\% of the reviewed platforms weight more than \SI{2.0}{\kilo\gram}.
    Considering only platforms with manipulators having at least 3 actuated \ac{DOF}, the lightest setup is based on a standard quadrotor and weights \SI{1.9}{\kilo\gram}~\cite{Fumagalli2014}. %, employing a standard quadrotor as aerial base.
    Instead, with only \SI{100}{\gram} more (total weight of \SI{2.0}{\kilo\gram}), we propose a novel overactuated platform, capable of independent pitching, with double the payload.

    Apart from the weight, compliance plays an important role in the contact stability during \ac{APhI}, as already shown in~\cite{suarez2016lightweight,bartelds2016compliant}.
    Despite its importance, the current state-of-the-art platforms still employ rigid-link delta manipulators, sometimes adding small spring elements at the end-effector~\cite{Fumagalli2014,tzoumanikas2020aerial}. This further increases the complexity and weight of the mechanical structure.
}
On the other hand, origami manufacturing allows for the lightweight construction of complex 3D structures through folding composites of rigid and flexible layers, generating links and joints with inherent flexibility~\cite{Rus2018}.
Specifically for delta robots, origami manufacturing facilitates ease of monolithic construction or miniaturization, such as for haptic user interfaces~\cite{Mintchev2019} or centimeter~\cite{Kalafat2021} and millimeter scale~\cite{McClintock2018} delta robots.

Despite the compliance and reduced weight of origami manipulators which make them well-suited for aerial applications, their use for aerial manipulation have remained mostly unexplored.
In~\cite{Kim2018c}, a one \ac{DOF} unarticulated origami arm was used as an extensible gripper on a \ac{MAV}, by storing the arm flat during take-off and extending it during flight.
%Mounting a three \ac{DOF} origami delta arm on a \ac{MAV} allows for a lightweight addition to improve utility and accuracy.

% \subsection*{Contributions}
In this work we describe the design and control of a novel small-size, lightweight, and overactuated \ac{AM}, representing a highly versatile platform for inspection tasks.
% The overactuation allows for efficient flight in a large range of different pitch angles as well as for fast force compensations when in physical interaction.
\EC{
    Its core elements are a tri-tiltrotor \ac{MAV} with 5 \acp{DOF} and an origami delta arm providing additional 3 \acp{DOF} (see Fig.~\ref{fig:voliroPlusOrigami}).
    The entire system has a take-off mass of \SI{2.0}{\kilo\gram} and its longest side spans a length of only \SI{56}{\centi\meter}.
    To the best of the authors' knowledge, this represents the lightest 8 \acp{DOF} \ac{AM} in the state-of-the-art.
    We demonstrate the use of an inherently compliant origami delta manipulator on an aerial robot, which allows for precise motion compensation tasks (as for rigid delta arms), while providing additional compliance during interaction.
    In particular, we first compute the exact delta kinematics, taking into account the non-idealities of the universal joints approximation.
    Then, we experimentally characterize the arm compliance and show how this affects the maximum force that the \ac{AM} can exert on the environment.
    Moreover we also discuss possible unwanted foldings of the origami joints (which we refer to as \emph{critical configurations}) and provide some insights on how to improve the prototype's robustness.
}
%
% \begin{itemize}
%         \item Mechanical design of a novel lightweight \ac{MAV}.
%         \item Implementation of a control scheme to exploit the pitch overactuation.
%         \item Mechanical design of the origami-based delta manipulator for \ac{APhI}.
%         \item Compliance characterization and control fo the origami-based manipulator.
% \end{itemize}
% The system will be validated through:
% \begin{itemize}
%         \item Free flight experiments to show independent translation and pitching.
%         \item Kinematic compensation in free flight with the origami delta.
%         \item Pushing on a surface with different compliances.
% \end{itemize}
\section{System design}
We design the system with the following goals in mind:
\begin{enumerate*}[label=(\roman*)]
    \item Small size and light weight platform, to allow operations in confined areas, while increasing safety when operating close to humans.
    \item Power efficiency for long flight times.
    \item Versatility and suitability for inspection tasks, including the ability to accurately touch an arbitrarily oriented surface at a desired location.
\end{enumerate*}
In order to achieve these goals, we design an overactuated platform augmented with an articulated end-effector. In our case, the overactuation has two main advantages: \begin{enumerate*}[label=(\roman*)]
    \item it allows instantaneous force compensation in the forward interaction direction, and
    \item it allows to hover and perform interaction tasks at different pitching angles.
\end{enumerate*}
An overview of the \ac{AM} is illustrated in Fig.~\ref{fig:voliroPlusOrigami}.

\subsection{Aerial platform}
We use the North-East-Down (NED) convention to describe the body frame $\frameB=\{\B O,\B\x,\B\y,\B\z\}$, fixed to the \ac{COM} of the \ac{MAV}, as well as the inertial world frame $\frameW=\{\W O,\W\x,\W\y,\W\z\}$, as illustrated in Fig.~\ref{fig:voliroPlusOrigami}.
The chosen multirotor configuration consists of two groups of coaxial rotors, which can tilt around the $\B\y$ axis, and a rear motor with a bidirectional propeller.
The propellers and tilt arms have complete control authority over the generated body torques, as well as forces along the $\B\z$ and $\B\x$ axes.
Since forces along $\B\y$ cannot be generated with the chosen configuration (they are always zero), the linear and angular dynamics in this direction are coupled.

The \ac{MAV} is intentionally built to fit inside small manholes and to operate in closed environments.
Without propeller guards it measures \SI{55}{\centi\meter} in length and \SI{56}{\centi\meter} in width.
The propellers of the two main rotor groups are 9"x4.7 while the rear 3D propeller measures 8"x4.5.
%, and the propeller guards protect them from small accidental collisions.
%
It weights \SI{1.7}{\kilo\gram} and can transport a maximum additional payload of \SI{1}{\kilo\gram}.
The main frame and all the other structural parts are printed with Nylon PA12 using a HP Multi Jet Fusion 3D printer, apart from motor arms and tail that are made of carbon tubes.
The biggest contribution on the weight comes from a \SI{5000}{\milli\ampere{}\hour} 4S battery (\SI{540}{\gram}), which experimentally reflected in around 18 minutes of flight time without additional payload.
Another experimental test with \SI{0.8}{\kilo\gram} of payload, for a total weight of \SI{2.5}{\kilo\gram}, resulted in 7 minutes of flight time.
A summary of the weight contributions of the different components is in table \ref{tab:weight}.
%
%The battery is placed under the body center and the \ac{COM} of the platform can be easily adjusted by shifting the battery forward or backward in the body $x$ axis.
\begin{table}[t]
    \begin{center}
        \EC{
            \begin{tabular}{|c|c|c|}
                \hline
                Part        & \ac{MAV} (\SI{}{\g}) & Delta (\SI{}{\g}) \\ \hline \hline
                Battery     & 540                  & -                 \\
                Motors      & 375                  & 135               \\
                Structure   & 345                  & 130               \\
                Electronics & 230                  & 25                \\
                Others      & 150                  & 10                \\
                Bumpers     & 60                   & -                 \\
                \hline
                \hline
                Total       & 1700                 & 300               \\
                \hline
            \end{tabular}
            \caption{Weight (in grams) of the \ac{AM} components.}
            \label{tab:weight}
        }
    \end{center}
    \vspace{-1em}
\end{table}
%
% The core of the body contains all the electronics.
%The servomotors are mounted in the center and the arms with the coaxial rotor groups are connected to them and supported by four ball bearings (two on each side).
%
% Also the spider clutches connecting the arms with the servos are made of TPU (Thermoplastic Polyurethane) to transmit the torque but at the same time providing some flexibility to axial forces.
%
\begin{table}[t]
    \begin{center}
        \EC{
            \begin{tabular}{|c|c|c|}
                \hline
                Component         & Name               & Qty. \\ \hline \hline
                Motor             & KDE2315XF-885      & 5    \\
                ESC               & Tekko32 F3         & 5    \\
                Origami servo     & Dynamixel XL-330   & 3    \\
                Tiltrotor servo   & Dynamixel XL-320   & 2    \\
                Buck Converter    & Henge 8A UBEC      & 2    \\
                Flight controller & Pixhawk Cube Grey  & 1    \\
                Battery           & Zop 4S 5000mAh     & 1    \\
                PDB               & Pixhawk 4 Mini PDB & 1    \\
                RC Receiver       & Jeti PPM Receiver  & 1    \\
                Onboard computer  & Raspberry Pi 4B    & 1    \\
                \hline
            \end{tabular}
        }
        \caption{Electrical components of the \ac{AM}.}
        \label{tab:electrical_components}
    \end{center}
    \vspace{-2em}
\end{table}
% \item Electric design.
A voltage buck converter provides \SI{5}{\volt} to power the Pixhawk flight controller, onboard computer and delta arm, while another provides \SI{7.5}{\volt} for the two Dynamixels XL-320 moving the arms.
%
% The other buck converter outputs a voltage of \SI{7.5}{\volt} that is transmitted to the two Dynamixels XL-320 moving the arms.
%
% The resolution of the Dynamixel is of $0.29^\circ$ and the stall torque produced is \SI{0.39}{Nm} at a stall current of \SI{1.1}{\ampere}.
%
% The flight controller receives commands from the remote controller or the onboard computer and sends signals to all the actuators.
Table \ref{tab:electrical_components} gives an overview of the platform's electrical components.
%
% The Dynamixels receive TTL signals from the Pixhawk through a UART serial port, to which pull-up resistors were added to achieve reliable communication with up to eight Dynamixels.
% %
% Data are transmitted asynchronously through the serial port at a baud rate of 1 MBaud using DYNAMIXEL Protocol 2.0.
% without payload needs a total of 19A, considering only the two main coaxial rotor groups, which should guarantee the flight time previously stated of 25 minutes.
%
\begin{figure}[t]
    \centering
    \includegraphics[width = 1\columnwidth]{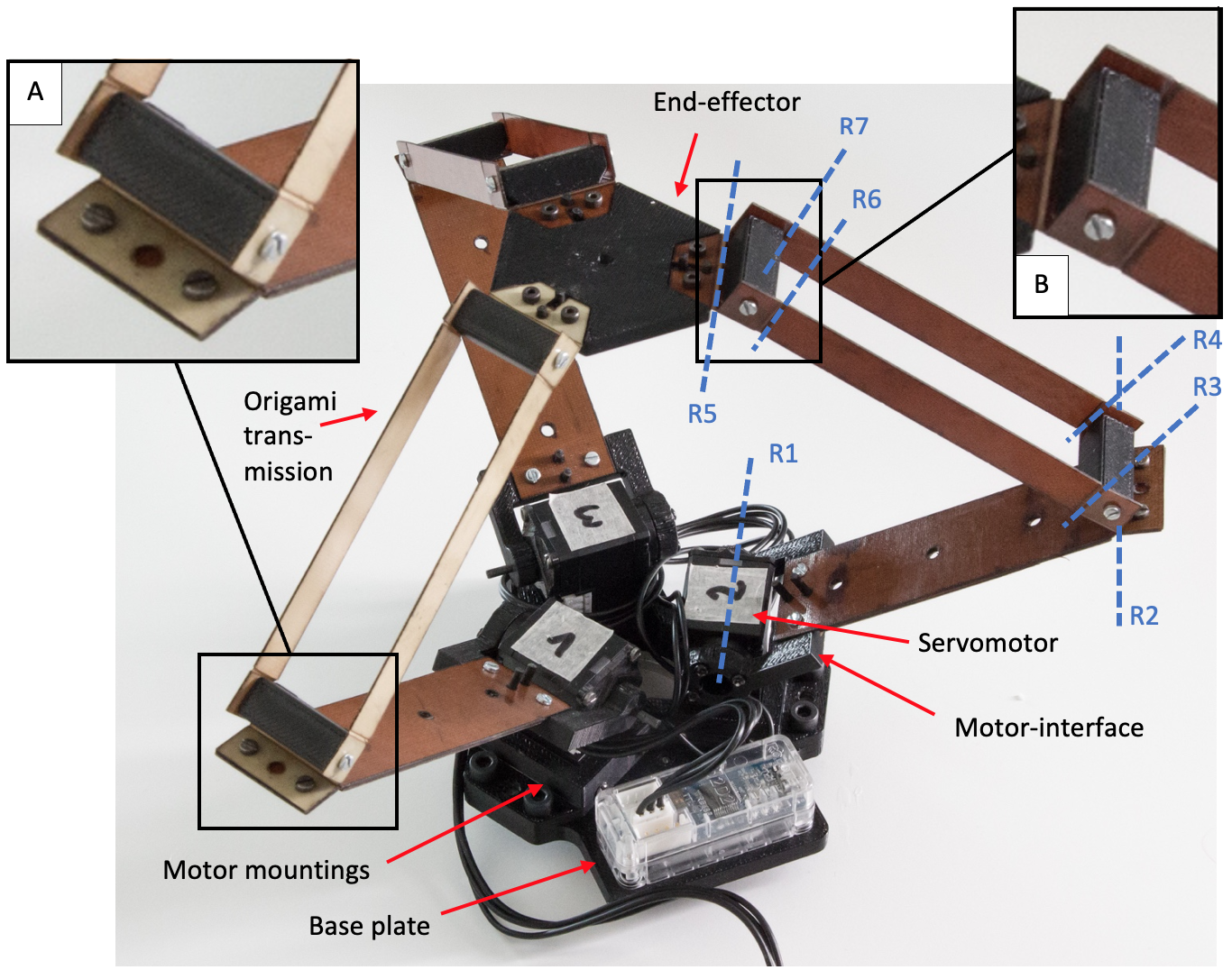}
    \caption{Design of the origami-based delta manipulator with the main components highlighted. The details (A) and (B) show the approximation of the universal joints using perpendicular revolute joints. The dashed lines indicate the rotation axes of each joint.}
    \label{fig:overview}
    \vspace{-1em}
\end{figure}

\subsection{Origami-based delta manipulator}
The design of the origami manipulator is shown in Fig.~\ref{fig:overview}.
Delta parallel arms consist of three identical legs connecting the moving end-effector plate to the fixed base platform.
Each of the legs is driven by a one \ac{DOF} rotary servomotor connected to the base platform.
The legs consist of proximal and distal limbs, the latter formed by parallel bars, creating a quadrilateral parallelogram-like structure.
This results in pure translation motions between base and end-effector~\cite{Rey1999}.

In ideal delta manipulators, the joints between the proximal limbs, the parallelogram linkages and the end-effector plate are universal joints.
For the origami manipulator, the universal joints are approximated using the solution proposed in~\cite{Mintchev2019,McClintock2018}. The side linkages of the parallelogram are folded upwards and an additional fold is added close to both ends of the linkages (Fig.~\ref{fig:overview}(A)).
This results in perpendicular revolute joints at the knee (R2, R3 and R4) and ankle (R5, R6 and R7) of the parallelogram, approximating the universal joints.
Unlike conventional delta robots, here the rotation axes do not coincide, resulting in a different kinematic model.
\EC{
    Even if new origami designs were proposed in order to remove the universal joint approximation~\cite{Kalafat2021}, we prefer to avoid having one monolitic arm structure.
    The origami limbs are attached to the 3D printed motor-interface and end-effector plate with screws and alignment pins, which allow easy replacement in case of breaking of an arm's element~\cite{Salerno2020}.
    Moreover we will derive the exact arm kinematics even in the presence of non-universal joints.
}
%
% The interface itself is screwed to the motor flange on one side and stabilized on the other side with the help of a bearing attached to the motor, creating joint R1.
%
% \begin{table}[t]
%     \begin{center}
%         \caption{Weight of the different parts included in the manipulator.}
%         \label{table:weights}
%         \begin{tabular}{| c | c |} \hline
%             Parts                     & Weight [g] \\ [0.5ex] \hline \hline
%             Motors + Cables           & 55.35      \\
%             Microcontroller           & 11.00      \\
%             MAV-Manipulator interface & 6.97       \\
%             Mounting structures       & 52.59      \\
%             Moving Parts              & 49.23      \\
%             Of which origami linkages & 30.60      \\
%             Others                    & 59.80      \\ \hline \hline
%             Total                     & 234.94     \\ \hline
%         \end{tabular}
%     \end{center}
% \end{table}
%
% Table \ref{table:weights} shows the weight of the different components of the manipulator.
%
% Most of the weight is located near the base of the manipulator, resulting in a reduced influence on the dynamics of the \ac{MAV} due to the manipulator motion.

%A major advantage of origami-based structures is the ability to use planar fabrication tools and methods to generate spatial mechanisms by folding along the joints.
%
The origami structures are made of a three layer, multi-material composite.
The top and bottom layers are made of fiberglass (FR-4-HF, 0.3mm), which provides the necessary stiffness. The middle Kapton layer with adhesive on both sides (DuPont Pyralux LF0111, 0.05mm) adds the necessary flexibility at the joints and bonds layers together.
Each layer is individually laser-cut with a $CO_2$ laser (Trotec Speedy 360), stacked, and then aligned by pins and holes.
The layers are then bonded together by the adhesive on the surfaces of the Kapton layer using a hydraulic heat press (Fontjine LabManual 300).
%
%The resulting composite is laser-cut again to remove the supports holding the rigid layers together during fabrication.
%
The upper and lower limbs are built separately and then screwed together at the knee joints, reducing material waste and simplifying the pattern.
%
% Additionally, it allows for faster and easier replacement if something breaks.
% %
% Nonetheless, it would be possible to fabricate each transmission or even the whole mechanism as one single structure.
%
The other structural parts were printed with ABS.
%using a Stratasys~F120.

The use of foldable joints instead of mechanical joints gives a compliant behavior to the delta robot.
This inherent flexibility depends not only on the design parameters (e.g., joint geometry and material properties), but also on the configuration assumed by the manipulator.
This allows adjusting the compliance of the manipulator to the requirements of the task.
For example, a soft configuration is conducive to safer interactions, while a stiffer configuration may be preferred to achieve greater accuracy and repeatability.
In Sec.~\ref{sec:experimental} we characterize the axial stiffness of the manipulator and show how it influences a simple interaction task.
%
%Nonetheless, it is also the reason for unwanted vibrations during rapid movements, as well as possible failure of the delta mechanism when the arms are widely extended, because the joints are insufficiently stiff to hold the configuration against the force of gravity and collapse.
%
%To adjust the stiffness of the system, different materials could be used or the structure could be modified by adding additional layers to the composite.
%
%The latter was implemented on the manipulator to reduce lower limb bending. %
%In the first design iteration, two additional fiberglass layers were laser cut and then simply screwed to the existing structure using the holes needed for manufacturing (Fig. \ref{fig:overview}).
%
%Later, the additional layers were bonded with the others during manufacture, saving the extra weight of the screws.
%\end{itemize}

\section{Control design}
\begin{figure}[t]
    \centering
    \includegraphics[width = 1\columnwidth]{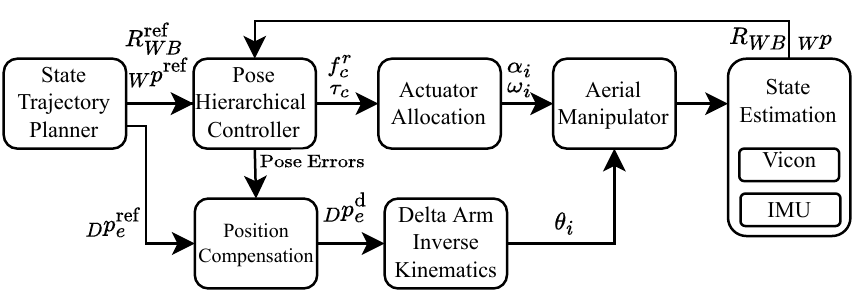}
    \caption{Control scheme of the \acl{AM}.}
    \label{fig:control_scheme}
    \vspace{-1em}
\end{figure}
In this section we present the control design of the \ac{AM}, which is schematically presented in Fig.~\ref{fig:control_scheme}.
First, we develop a pose control law for the overactuated \ac{MAV} modifying the geometric controller for quadrotors in \cite{leeSE3}.
Secondly, we derive the \ac{IK} for the origami delta arm, coupling it to the \ac{MAV}'s pose.
\subsection{Aerial platform pose control}
Consider the inertial world frame $\frameW$ and the body frame $\frameB$ attached to the \ac{MAV} in its \ac{COM}.
We define $\W\pos$ as the position of the body frame's origin in $\frameW$ and $\rotMatBW \in \mathrm{SO}(3)$ as the rotation matrix from $\frameW$ to $\frameB$.
The position and attitude dynamics of the \ac{MAV} are then given by
\begin{subequations}
    \label{drone_model}
    \begin{IEEEeqnarray} {ll}
        \mass\left(\B{\acc}+\B{\angVel}\times\B{\vel}\right) = \B\gravityForce + \B\forceCommand\\
        \inertia\B\angAcc + \B\angVel\times \inertia\B{\angVel} = \torqueCommand,
    \end{IEEEeqnarray}
\end{subequations}
where $\mass \in \nR{}$ is the total mass of the platform, $\inertia \in \nR{3\times3}$ is the inertia matrix in $\frameB$, $\gravityForce \in \nR{3}$ is the gravity force vector, $\vel,\angVel \in \nR{3}$ are the platform's linear and angular velocity, and $\forceCommand,\torqueCommand \in \nR{3}$ are the force and torque commands, respectively.
Since the system cannot produce instantaneous thrust along its body $\B\y$ axis, we employ a cascaded control structure with an outer loop position controller and an inner loop attitude controller.
%a hierarchical controller between position and attitude is necessary.
%
%The control hierarchy still addresses the full actuation in the pitching dynamics.
%
Consider the position and velocity errors of the linear dynamics as
\begin{subequations}
    \label{eq:pos_errors}
    \begin{IEEEeqnarray} {ll}
        \B\positionError &=  \rotMatBW \left(\W\pos\refer - \W\pos\right)\\
        \B\velocityError &= \rotMatBW\W\vel\refer - \B\vel,
    \end{IEEEeqnarray}
\end{subequations}
where $\positionError,\velocityError \in \nR{3}$ are the position and velocity errors, respectively, and the quantities $(\cdot)\refer$ are generated by a suitable trajectory planner.
%
%From \eqref{pos_errors} the force control vector $\forceCommand \in \nR{3}$ can be computed as
Based on \eqref{eq:pos_errors} we define the Proportional-Derivative (PD) control law as
\begin{multline}
    % \begin{split}
    \B\forceCommand = \kpPos \B\positionError + \kdPos \B\velocityError - \B\gravityForce +\\+ \mass \left(\rotMatBW\W\acc\refer+\B{\angVel}\times\B{\vel}\right).\label{eq:forcecmd}
    % \end{split}
\end{multline}
%Note that $\B\forceCommand$ is a vector in $\nR{3}$, but the platform cannot exert forces in the ${}_B\vec{y}$ body axis.
For the attitude control loop, consider the reference orientation given by $\rotMatWB\refer = \left[ \B{\vec{x}}\refer \vSpace \B{\vec{y}}\refer \vSpace \B{\vec{z}}\refer \right]\in\mathrm{SO}(3)$ that contains the pitch and yaw angle references of $\frameB$ w.r.t. $\frameW$ (note that, given the platform's actuation, only yaw and pitch can be tracked individually).

We then construct the attitude-loop target orientation, denoted by $\rotMatWB\des\in\mathrm{SO}(3)$, as follows.
%Given the force command in \eqref{eq:forcecmd} and a reference orientation given by $\rotMatWB\refer\in\mathrm{SO}(3)$, we construct the attitude-loop target orientation, denoted by $\rotMatWB\des\in\mathrm{SO}(3)$ .
We first define a new command vector $\B\underForceCommand = \left[ 0 \vSpace \B{}\forceCommandi{y} \vSpace \B{}\forceCommandi{z} \right]\transpose$, from which the commanded force along $\B x$ has been removed. %, since the platform can exert it instantaneously and it does not need to be a reference for the attitude controller.
We then rotate it into $\frameW$ and compute the desired body-frame z-axis, expressed in the world frame, as $\B{\vec{z}}\des \coloneqq \frac{\W\underForceCommand}{\norm{\W\underForceCommand}}$.
%
%For the attitude control loop, consider the reference rotation matrix $\rotMatWB\refer = \left[ \B{\vec{x}}\refer \vSpace \B{\vec{y}}\refer \vSpace \B{\vec{z}}\refer \right]$ that contains the pitch and yaw angle references of $\frameB$ w.r.t. $\frameW$ (note that, given the platform actuation, only yaw and pitch can be tracked individually).
%
Lastly, we compute $\B{\y}\des = \frac{\B {\z}\des\times\B{\x}\refer}{\norm{\B{\z}\des\times\B{\x}\refer}}$ and obtain the desired rotation matrix for the inner attitude control loop as
\begin{equation}
    \rotMatWB\des = \matrix{ \B{\x}\refer & \B{\y}\des & \frac{\B{\x}\refer\times\B{\y}\des}{\norm{\B{\x}\refer\times\B{\y}\des}} } .
    \label{eq:desRot}
\end{equation}
Note how the desired rotation matrix in \eqref{eq:desRot} preserves the reference pitch and yaw angles, while exploiting the roll dynamics to pursue the position tracking task.
We now define the inner attitude loop control errors as
\begin{subequations}
    \label{ang_errors}
    \begin{IEEEeqnarray} {ll}
        \B\angError &= \frac{1}{2} \left[ \rotMatWB\des{}\transpose \rotMatWB - \rotMatWB\transpose \rotMatWB\des \right]^\vee, \\
        \B\angVelError &= \B\angVel - \rotMatWB \W\angVel\refer,
    \end{IEEEeqnarray}
\end{subequations}
where $(\cdot)^\vee:\mathfrak{so}{(3)}\rightarrow\nR{3}$ is the vee operator which transforms a skew-symmetric matrix to a vector.
Then, the control torque command $\B\torqueCommand \in \nR{3}$ can be computed as
\begin{equation}
    \begin{split}
        \B\torqueCommand &= \kpAng \angError + \kdAng \angVelError + \angVel\times \inertia \angVel + \\
        & - \inertia\left[\angVel\times\rotMatWB\transpose \rotMatWB\des\angVel\refer  - \rotMatWB\transpose \rotMatWB\des\angAcc\refer\right]
    \end{split}
\end{equation}
where $\kpAng,\ \kdAng \in \nR{3\times 3}$ are diagonal and positive gain matrices. % and $\skewSym{\cdot}:\nR{3}\rightarrow\mathfrak{so}(3)$ is the skew-symmetric operator.
All quantities are expressed in $\frameB$ and the $\B(\cdot)$ subscript has been omitted for brevity.
% \begin{figure}[t]
%     \centering
%     \includegraphics[width=1\columnwidth]{convention.png}
%     \caption{Convention for the overactuated \ac{MAV}.}
%     \label{pics:convention}
% \end{figure}
\subsubsection*{Actuator allocation}
In order to compute the actuator commands from the force and torque control commands, we compute the \emph{actuator allocation} as follows.
We define $T_{12}$ and $ T_{34}$ as the thrusts produced by the respective motor groups, and $\alpha_0$ and $\alpha_1$ as the tilt angles of the two arms, as showed in Fig.~\ref{fig:voliroPlusOrigami}. Furthermore, $l_{1}$ is the distance from the center of the two frontal motor groups to the origin of $\frameB$, and $l_{2}$ the distance from the center of the tail motor to the to the origin of $\frameB$.
Considering the \ac{MAV} geometry, the actuator allocation is given by the equations
%The moment produced by the thrust coefficient $k_{d}$ of the rear motor, multiplied with the thrust of the rear motor, is ignored into the allocation since it can be easily be handled by the controller. \\
\begin{equation}
    \begin{bmatrix}
        \reducedForceCommand \\
        \torqueCommand
    \end{bmatrix} =
    \underbrace{\begin{bmatrix}
            1   & 0   & -1  & 0    & 0    \\
            0   & -1  & 0   & -1   & -1   \\
            0   & l_1 & 0   & -l_1 & 0    \\
            0   & 0   & 0   & 0    & -l_2 \\
            l_1 & 0   & l_1 & 0    & -k_d \\
        \end{bmatrix}}_{\bm{A}}
    \underbrace{\begin{bmatrix}
            T_{12}\sin(\alpha_1) \\
            T_{12}\cos(\alpha_1) \\
            T_{34}\sin(\alpha_0) \\
            T_{34}\cos(\alpha_0) \\
            T_5
        \end{bmatrix}}_{\inputSin} ,
    \label{allocation}
\end{equation}
with $\reducedForceCommand = \left[ \B{}\forceCommandi{x} \vSpace \B{}\forceCommandi{z} \right]\transpose$ containing only command forces along $\B\x$ and $\B\z$.
% , and the input vector $\inputSin \in \nR{5}$
% \begin{equation}
%     \inputSin =
%     \begin{bmatrix}
%         T_{12}\sin(\alpha_1) \\
%         T_{12}\cos(\alpha_1) \\
%         T_{34}\sin(\alpha_0) \\
%         T_{34}\cos(\alpha_0) \\
%         T_5
%     \end{bmatrix} .
% \end{equation}
Given a set of control forces and torques (i.e. \emph{wrench}), we compute the input vector $\inputSin$ as
\begin{align}
    \inputSin=\bm{A}^{-1}\begin{bmatrix}
        \reducedForceCommand \\
        \torqueCommand
    \end{bmatrix},
\end{align}
and solve for the individual actuator commands:
%From \eqref{allocation}, the thrust commands and the tilt angles can be found as
\begin{equation}
    \begin{gathered}
        T_{12} = \sqrt{u_0^2 + u_1^2};\quad
        T_{34} = \sqrt{u_2^2 + u_3^2}; \quad T_5 = u_5\\
        \alpha_0 = \atanTwo(u_1,u_2);\quad
        \alpha_1 = \atanTwo(u_3,u_4).
    \end{gathered}
    \label{thrusts_tilts}
\end{equation}
From there the rotational speeds of the motors can be calculated by using the motor coefficients $k_{f}$, $k_{f,\text{rear}} \in \nR{}$ and assuming a quadratic relationship
\begin{equation}
    \begin{gathered}
        \omega_{1} = \omega_{2} = \sqrt{\dfrac{T_{12}}{2k_{f}}};\quad
        \omega_{3} = \omega_{4} = \sqrt{\dfrac{T_{34}}{2k_{f}}}\\
        \omega_{5} = \sign(T_5)\sqrt{\dfrac{\norm{T_{5}}}{k_{f, \text{rear}}}} .
    \end{gathered}
    \label{motor_relation}
\end{equation}
Note that the sign of the rear thrust $T_5$ must be specifically taken into account, since the rear motor is bidirectional.

\EC{
    Also, with $k_{f} = \SI{8.1e-6}{\newton\second\squared}$, $k_{f,\text{rear}} = \SI{4.05e-6}{\newton\second\squared}$ and maximum rotor speed $\omega_{\text{max}} = \SI{1143}{\radian}$, the platform achieves a maximum total thrust of $T_{\text{max}} = \SI{4.85}{\kilogram}$, which corresponds to a thrust-to-weight ratio of $\SI{2.43}{}$, considering the full \ac{AM}.
    This relatively high ratio allowed never entering into actuator saturation in the proposed experiments.
}

\subsection{Origami-based delta manipulator inverse kinematics}
\begin{figure}[t]
    \centering
    \includegraphics[width = 0.9\columnwidth]{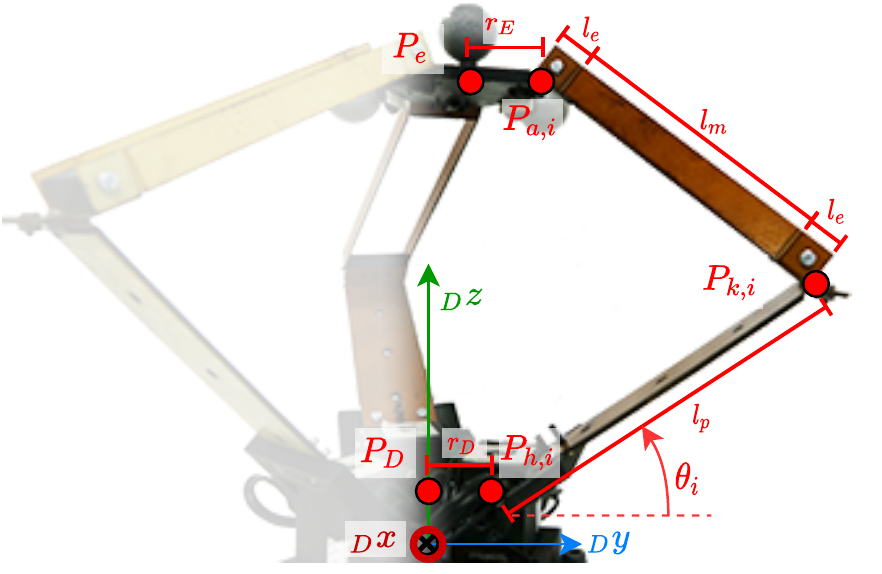}
    \caption{Schematics of the origami arm.}
    \vspace{-1em}
    \label{fig:schematics}
\end{figure}
In this section we describe an \ac{IK} approach to find the manipulator joint angles $\deltaJoint$ as function of the end-effector target position $\D\eefPoseDes$. We express the position of the end-effector frame $\frameE=\{\E O,\E\x,\E\y,\E\z\}$ in the delta frame $\frameD=\{\D O,\D\x,\D\y,\D\z\}$, which is fixed to the base plate of the arm, as in Fig.~\ref{fig:schematics}.
%In order for the delta manipulator end-effector to reach a desired target position $\D\eefPoseDes$, the corresponding angles for the actuated joints $\deltaJoint$ are needed.
We exploit the solution for conventional delta robots \cite{WilliamsII2016} with some adjustments to account for the kinematic differences of the origami adaptation.

Consider the geometric description of the origami delta arm in Fig.~\ref{fig:schematics}.
The solution for a conventional delta robot is given by finding the intersection point $P_{k,i}$ of a circle around the hip joint $P_{h,i}$ with radius $l_p$ and a sphere around the ankle joint $P_{a,i}$ with radius $l_d = l_m + 2 l_e$.
However, since in the origami design the universal joints do not have aligned rotation axes, the length of the distal link $l_d$ does not remain constant.
Therefore, we replace $l_d$ with an expression $l_{d,i}(\D\eefPose)$, which depends on the dimensions $l_m$ and $l_e$ as before, but is a function of the end-effector position as well.
%which consists of $l_m$ and $l_e$ and depends on the parallelogram angle.
%Due to the horizontal and vertical joints in the upper leg not aligning in the origami delta manipulator, its reachable workspace changes from a sphere to an oval shape. To account for this change in kinematics, the upper leg length $l_u$ needs to be replaced with an expression $l_{u,i}$ which consists of $l_m$ and $l_e$ and depends on the parallelogram angle.

The kinematic relation for a generic leg $i=\{1,2,3\}$ is
\begin{equation}
    \label{link_constraint}
    \norm{\lv_{d,i}}^2 = \norm{\D\eefPose + \D \Pv_{ai} - \D\Pv_{hi} - \D \lv_{ki-hi}(\deltaJoint)}^2,
\end{equation}
with $\D\lv_{ki-hi} = \left[ 0 \vSpace l_p\cos{\deltaJoint} \vSpace l_p \sin{\deltaJoint} \right]\transpose$ depending on the joint angle.
%,$\Pv_{h,i} = \left[ 0 \vSpace r_D \vSpace 0 \right]\transpose$, and $\Pv_{a,i} = \left[ 0 \vSpace r_E \vSpace 0 \right]\transpose$.
% \begin{equation}
%     \D \Pv_{h,i} =
%     \begin{bmatrix}
%         0 \\
%         r_D \\
%         0
%     \end{bmatrix};\quad \D \lv_{p,i} = 
%     \begin{bmatrix}
%         0 \\
%         l_p \cos{\theta_i} \\
%         l_p \sin{\theta_i}
%     \end{bmatrix}
% \end{equation}
From geometric considerations on the distal link parallelogram, we compute its true length
%
% from
% \begin{equation}
%     \lv_{d,i}(\D\eefPose) = 
%     \D\eefPose - 
%     \begin{bmatrix}
%         0 \\
%         l_p \cos{\deltaJoint} + r_{DE} \\
%         l_p \sin{\deltaJoint}
%     \end{bmatrix} ,
%     % a = r_D - r_E
% \end{equation}
as
\begin{equation}
    l_{d,i}(\D\eefPose) = \sqrt{p_{e,\parallel}^2 + \left(\sqrt{l_m^2 - p_{e,\parallel}^2} + 2l_e\right)^2} ,
    \label{distal_len}
\end{equation}
with $p_{e,\parallel}$ the component of $\D\eefPose$ parallel to the $P_{hi}$ joint axis.
% \begin{equation}
%     \label{ankle_transform}
%     \D \Pv_{a,i} =
%     \begin{bmatrix}
%         0 \\
%         r_E \\
%         0 \\
%     \end{bmatrix}
% \end{equation}
%where in \eqref{ankle_transform} we make use of the constraint that the end-effector platform of a delta robot always remains parallel to the base and is only allowed translational motion.
% To solve for the variable $\deltaJoint$ the constraint on the distal leg length is used
% % \begin{equation}
% %     \label{quadratic_equation}
% %     l_{d,i} = \norm{\D \lv_{d,i}} = 
% %     \norm{\D \Pv_{E} + \D \Pv_{a,i} - \D \Pv_{h,i} - \D \lv_{p,i}},
% % \end{equation}
% with the origami distal leg expression
% \begin{equation}
%     l_{d,i} = \sqrt{\D p_{E,x}^2 + \left(\sqrt{l_m^2 - \D p_{E,x}^2} + 2l_e\right)^2}
% \end{equation}
% By squaring both sides of \eqref{quadratic_equation} we finally get the equation:
Then, by combining \eqref{link_constraint} and \eqref{distal_len}, we get
\begin{subequations}
    \begin{IEEEeqnarray} {ll}
        0   &= E_i \cos{\theta_i} + F_i \sin{\theta_i} + G_i,\\
        E_i &= 2 l_p (r_{DE} - \D p_{E,y}), \\
        F_i &= - 2\D p_{E,z} l_p, \\
        G_i &= \D p_{E,x}^2 + \D p_{E,y}^2 + \D p_{E,z}^2 + r_{DE}^2 +\nonumber \\
        &+ l_p^2 - 2 \D p_{E,y} r_{DE} - l_{d,i}^2,
    \end{IEEEeqnarray}
    \label{eq:origami}
\end{subequations}
with $r_{DE} = r_D - r_E$.
%$\theta_1$ is finally solved for by applying the tangent half-angle substitution and converting \eqref{squared_leg_contraint_equation} to a quadratic equation:
%\begin{equation}
%        t_i \coloneqq \tan{\frac{\theta_i}{2}} \quad
%        \cos{\theta_i} = \frac{1 - t_i^2}{1 + t_i^2} \quad
%        \sin{\theta_i} = \frac{2t_i}{1 + t_i^2}
%\end{equation}
%\begin{equation}
%    (G_i - E_i)t_i^2 + (2F_i)t_i + (G_i + E_i) = 0
%\end{equation}
%By applying the tangent half-angle substitution in \eqref{squared_leg_contraint_equation}, the desired joint angle $\deltaJoint$ can be finally found as
Finally, in order to compute the desired joint angles $\deltaJoint$ for the end-effector to reach a target position $\D\eefPoseDes$, we set $\D\eefPose=\D\eefPoseDes$ and solve \eqref{eq:origami} as\footnote{We employed the tangent half-angle substitution to solve this equation.}
% \begin{equation}
%     \deltaJoint = 2 \tan^{-1}(t_i),
% \end{equation}
\begin{equation}
    \deltaJoint(\D\eefPoseDes) = 2 \tan^{-1}\left(\frac{-F_i + \sqrt{E_i^2 + F_i^2 - G_i^2}}{G_i - E_i}\right).
\end{equation}
The desired joint angles are then tracked by the servomotors' integrated PID controllers. %
\begin{figure}[t]
    \centering
    \includegraphics[width=0.8\columnwidth]{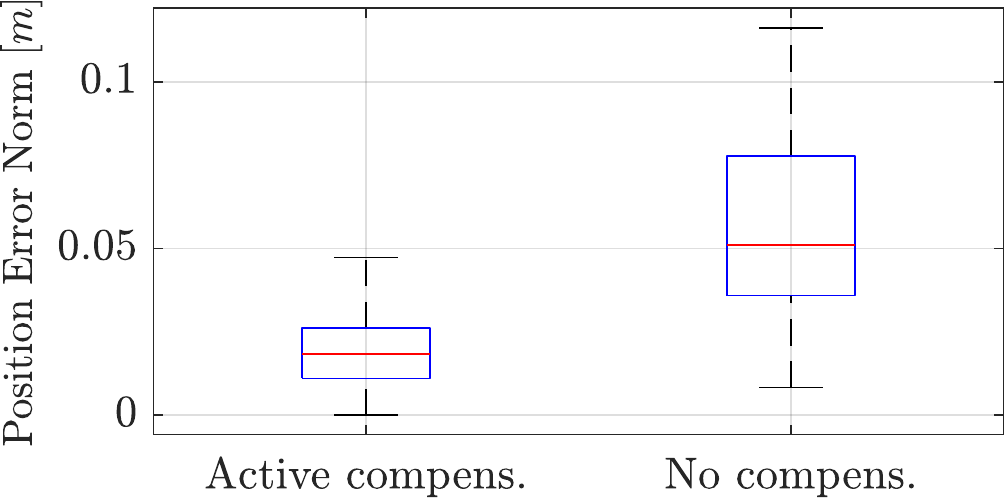}
    \caption{Position error norm at the end-effector with and without active compensation. The two distributions are positively skewed with medians \SI{0.02}{\meter} and \SI{0.05}{\meter}, respectively.}
    \label{pics:pos_tracking_errors}
    \vspace{-1em}
\end{figure}
\subsection{Kinematic coupling law}
To control both the \ac{MAV} body and the delta arm, we couple the two frames $\frameB$ and $\frameE$ kinematically. This aims to compensate any oscillations occurring in the body pose tracking error $\B\positionError$.
To this end, we adapt the end-effector position reference $\D\eefPoseRef$ according to $\B\positionError$, generating the instantaneous end-effector target $\D\eefPoseDes$.
%In particular, consider commanding a reference trajectory $\W\eefPoseRef$ to the \ac{AM}'s end-effector, distributed between the \ac{MAV} body reference $\left( \W\pos\refer, \rotMatWB\refer \right)$ and a delta arm reference $\D\eefPoseRef$.
%
%Because of possible oscillations of the aerial base due to external disturbances, the planned trajectory of the end-effector would also be affected.
%
%For this reason, we compensate the end-effector reference by commanding a desired position
\begin{equation}
    \D\eefPoseDes = \D\eefPoseRef + \rotMatDB\B\positionError + \rotMatDB\rotMatWB\refer{}\transpose \rotMatWB \B \pos_{BD} ,
\end{equation}
where $\B \pos_{BD} \in \nR{3}$ is the distance between the origins of the $\frameD$ and $\frameB$ expressed in $\frameB$ and $\rotMatWB\refer{}\transpose \rotMatWB$ accounts for rotational errors.
The instantaneous target $\D\eefPoseDes$ is then used to compute the desired joint angles $\theta_i$.

\section{Experimental results}\label{sec:experimental}
% \subsection{Mini-voliro only trajectory tracking}
In this section, we focus specifically on three aspects: \begin{enumerate*}[label=(\roman*)]
    \item The end-effector position tracking performance during free flight,
    \item a characterization of the origami manipulator stiffness depending on its configuration, and
    \item the system characteristics during interaction, particularly the achievable interaction forces with different manipulator stiffness.
\end{enumerate*}
\subsection{Manipulator kinematic compensation}
In this experiment, we evaluate the positioning accuracy of the \ac{AM}'s end-effector.
%
% Note that the position errors at the end-effector need special care since they are not only affected by the body position, but also its orientation error.
%
To perform this analysis, we command a constant end-effector reference pose and we track it only using the \ac{MAV}, with the origami arm in a fixed configuration.
After recording a sufficient number of samples, the delta arm is actively commanded to compensate for the floating base displacements.
The end-effector position is obtained from the arm's forward kinematics. %
The performances in the two scenarios are in Fig.~\ref{pics:pos_tracking_errors}.
The median and interquartile range of the tracking error's norm is more than halved in active compensation with respect to the fixed arm case.
%
% The same goes for the interquartile range, reducing oscillations at the arm's tip.
%
% \EC{
%     Also, despite the flexibility on the origami structure and the fast motions commanded, no unwanted oscillations were noticed in the arm's links during the experiment.
% }

\subsection{Origami manipulator stiffness characterization}
\begin{figure} [t]
    \centering
    \begin{subfigure}{0.45\columnwidth}
        \includegraphics[width=\columnwidth]{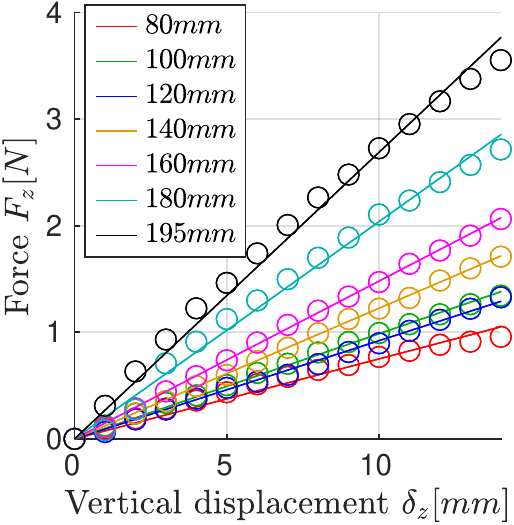}
        \caption{Linear spring model fits at different heights $\D p_{E,z}$.}
        \label{pics:compliance_data}
    \end{subfigure}
    \hfill
    \begin{subfigure}{0.49\columnwidth}
        \includegraphics[width=1\columnwidth]{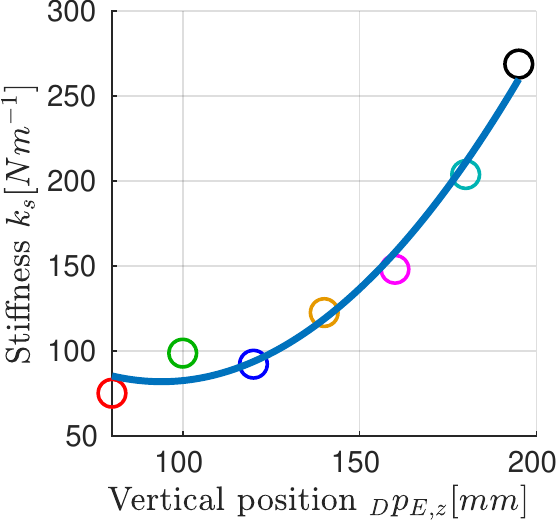}
        \caption{Stiffness interpolation over height.}
        \label{pics:stiffness_factor}
    \end{subfigure}
    \caption{Origami delta arm stiffness identification.}
    \vspace{-1em}
\end{figure}
Here we model the manipulator's stiffness in different positions of the end-effector frame $\frameE$ with respect to the delta base frame $\frameD$.
Specifically, the origin of $\frameE$ was always kept along the vertical direction $\D\z$. %along the center axis.
These measurements were taken by attaching the manipulator on a load cell and pressing down the end-effector plate in one millimeter increments along the $\D\z$ direction. % while measuring the applied force.\\
Both end-effector position displacements $\delta_z \in \nR{}$ and push force $F_z \in \nR{}$ were recorded for each chosen end-effector height value $\D p_{E,z}$.
In the end, a linear spring model was fitted for different height values as $F_z = \Kstiffness\delta_z$, with $\Kstiffness(\D p_{E,z})$ the estimated end-effector stiffness. The different stiffness fittings are visible in Fig.~\ref{pics:compliance_data}.
In particular, different data points have been collected at heights in the range $\D p_{E,z} \in \left[80, \vSpace 195 \right] \si{\milli\meter}$, interpolating the resulting stiffness with a second order polynomial $\Kstiffness(\D p_{E,z}) = c_0 + c_1 \D p_{E,z} + c_2 \D p_{E,z}^2 $, where $c_0$,$c_1$,$c_2 \in \nR{}$ are the identified coefficients.
We choose a second-order curve to balance complexity and fitting error.
The resulting interpolation is shown in Fig.~\ref{pics:stiffness_factor}, with a stiffness change in the range $\Kstiffness \in \left[80 \vSpace 290 \right] \si{\newton\per\meter}$.
Note that the stiffness is greater when the arm is fully extended, whereas it is most compliant with the arm retracted.
Having a mechanically variable stiffness arm represents an advantage when it is not possible to implement a software impedance control action.
For instance, this is the case when the mathematical model of the \ac{MAV} is not known with enough precision for an impedance controller, or when only position control is available without an external \ac{FT} sensor or estimator to implement an admittance control scheme.
\EC{
    Note that the stiffness characterization is performed considering a centered end-effector since we assume to only interact with the environment in this condition.
    A complete characterization, although possible, would require much more experimental data and is beyond the scope of this work.
    Here we aim at providing preliminary results on how the compliance influences the interaction contact forces, whereas a full exploitation will be addressed in future works.
}
%A linear spring model $F = k*dz$ was fit to the first 15mm of data for each recorded manipulator position. A quadratic fit was then used to interpolate in between the measured data points.
%   The resulting model showed a possible stiffness coefficient $k$ in the range from 82 $\frac{N}{m}$ up to 317 $\frac{N}{m}$.
%
\subsection{Physical interaction}
\begin{figure}[t]
    \centering
    \includegraphics[width=1\columnwidth]{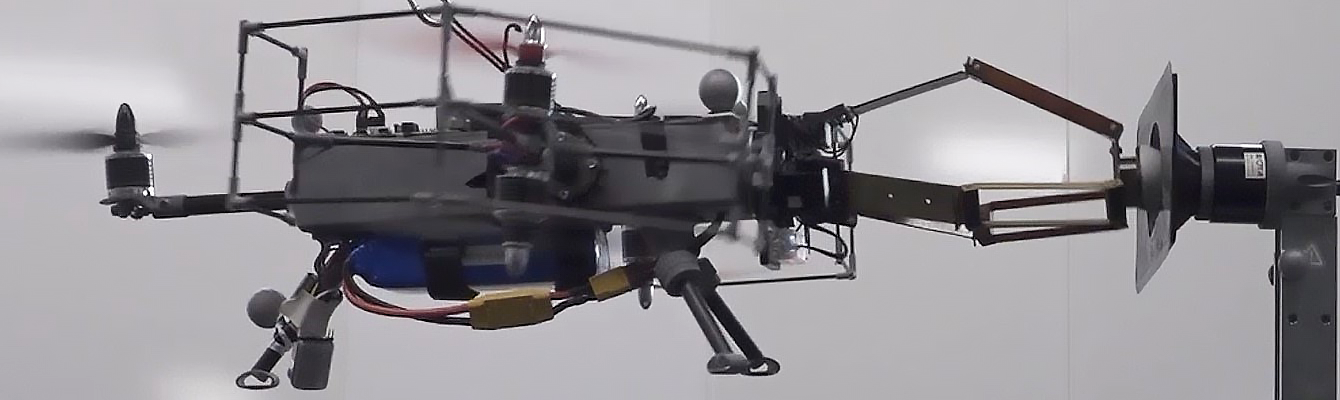}
    \caption{\ac{APhI} experiment with the \ac{AM} pushing on a surface connected to a \ac{FT} sensor.}\label{fig:force_experiment_pic}
    \vspace{-1.0em}
\end{figure}

\begin{figure}[t]
    \centering
    \begin{subfigure}{1.0\columnwidth}
        \includegraphics[width=\columnwidth]{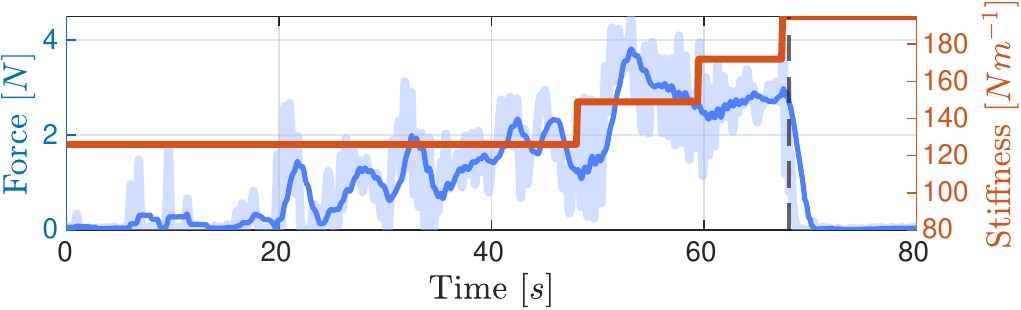}
        \caption{Interaction with increasing stiffness.}
        \label{pics:interaction_max_stiff}
    \end{subfigure}
    \hfill
    \begin{subfigure}{1.0\columnwidth}
        \includegraphics[width=\columnwidth]{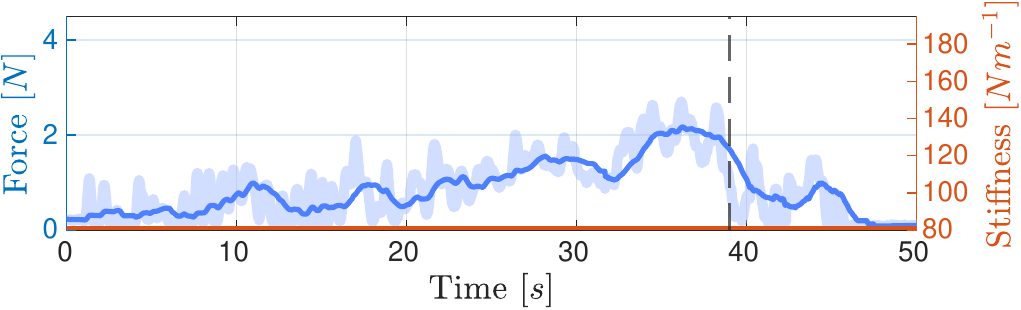}
        \caption{Interaction with constant low stiffness.}
        \label{pics:interaction_min_stiff}
    \end{subfigure}
    \caption{Interaction force (blue) generated with different stiffness configurations (red). The dashed vertical line represents the time instant of the origami critical folding. The transparent blu shadow represents the unfiltered force measurements. The solid blue line is a filtered version for the graph's clarity.}
    \label{pics:physical_interaction_plot}
    % \vspace{-0.8em}
\end{figure}

To study the different behaviors and exerted forces depending on the commanded stiffness configuration of the manipulator, a physical interaction experiment was conducted as shown in Fig.~\ref{fig:force_experiment_pic}.
%
%An interaction surface attached to a load cell was set up at approximately \SI{1.5}{\meter} in height.
%
We commanded the \ac{AM} to approach a surface and push against it with the origami manipulator configured at different stiffness levels.
The surface was connected to a \ac{FT} sensor to provide ground truth data of the pushing forces.
%
%In order to generate increasing pushing forces, the \ac{AM} was commanded a position reference such as to enter in interaction with the surface.
%
Once in contact, the origami arm's stiffness $\Kstiffness$ was increased up to \SI{290}{\newton\per\meter}.
This resulted in a subsequent increase of the pushing force, as in Fig.~\ref{pics:interaction_max_stiff}.
The origami manipulator was able to sustain a peak force of $\SI{4}{\newton}$ before the structure folded into a critical configuration.
Similarly, another experiment was conducted with the lowest stiffness allowed by the manipulator, while pushing further with the aerial base, in Fig.~\ref{pics:interaction_min_stiff}.
Here, the \ac{AM} was exerting a force of $\SI{2}{\newton}$ when the critical folding occurred.
%Using the lowest stiffness configuration the manipulator interacted softly and with maximum exerted forces of approximately \SI{1.5}{\newton}. At the highest stiffness configuration the exerted forces increased up to \SI{4}{\newton} and the observed interaction became more rigid.
%
\begin{figure} [tbh]
    \centering
    \begin{subfigure}{0.49\columnwidth}
        \includegraphics[width=\columnwidth]{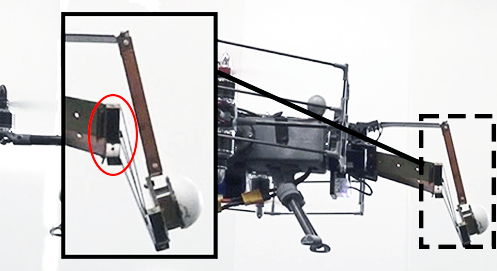}
        % \caption{Knee critical folding.}
        \label{pics:lateral_fold}
    \end{subfigure}
    \hfill
    \begin{subfigure}{0.49\columnwidth}
        \includegraphics[width=\textwidth]{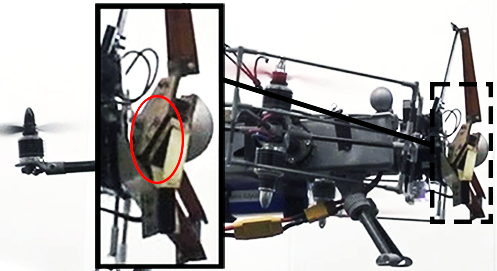}
        % \caption{Ankle critical folding.}
        \label{pics:inside_fold}
    \end{subfigure}
    \caption{Knee (left) and ankle (right) critical foldings.}
    \label{pics:critical_configurations}
    \vspace{-1.0em}
\end{figure}

The \emph{critical folding configurations} are phenomenons due to an unnatural bending of the origami joints, from which the \ac{AM} can't autonomously recover, as in Fig.~\ref{pics:critical_configurations}.
We refer to \emph{knee} or \emph{ankle critical folding} as the ones caused by the bending of one of the knee or ankle joints, respectively.
%We refer to \emph{knee critical folding} as the one caused by an outward bending of one of the knee joints.
%
The former is likely to happen when pushing too strongly on the fully extended origami arm, i.e., in a high-stiffness configuration.
%
%We then refer to \emph{ankle critical folding} as the one caused by an inward bending of one of the ankle joints, represented in Fig.~\ref{pics:inside_fold}.
%
The latter happens when pushing too much on the retracted origami arm, in a low-stiffness configuration.
Manipulator configurations at the center of the stiffness spectrum were generally less prone to fold into critical configurations.
Identifying these particular configurations is of primary importance to describe the arm's feasible workspace and exerted forces, to avoid criticalities in more complex tasks, where reliability plays a fundamental role.
\EC{
    In particular, future designs of the origami manipulator will feature mechanic stoppers at joints $R5$ to prevent the \emph{ankle critical folding}, while increasing the robustness of joints $R3$ and $R4$ will prevent the \emph{knee critical folding}.
}
%What primarily limited the maximum force were critical folding configurations under high pressure from which the manipulator could hardly recover from. At compact low-stiffness configurations the manipulator would snap to the inside at approximately \SI{1.5}{\newton}. On the other hand in the more extended positions at high stiffness the manipulator was prone to buckling downwards at forces of approximately \SI{4}{\newton}. This later effect is caused by the manipulator being slightly bent downwards due to gravity. Therefore the forces pushing on the manipulator were not acting well centered. Manipulator positions at the center of the stiffness spectrum were generally more stable and less prone to fold into critical configurations.

\section{CONCLUSIONS}
We realized a small and lightweight \ac{AM} for inspection purposes.
We first described the construction process of both the aerial platform and the origami arm and how a very low weight can be reached with a careful choice of design and building materials.
We then derived a pose controller for the body and an \ac{IK} controller for the manipulator, coupling them to achieve accurate pose tracking of the end-effector.
%The resulting controller can deal with both the overactuation of the platform and the non-idealities of an origami delta arm. % with respect to the standard counterpart.
%
We showed how the addition of the origami manipulator increases the end-effector tracking performance in free-flight and how the manipulator compliance can be adjusted during \ac{APhI}, affecting the generated interaction forces.
We validated the use of inherently compliant manipulators as opposed to the rigid counterparts with additional spring elements, which would increase the system's weight and complexity.
%
%also highlighted the advantages of having such a flexible arm, 
In the end, we also analyzed its limitations when it comes to undesirable arm foldings.
We believe that in future work, adjusting the manipulator compliance in such a high range will be a key element in more complex interaction scenarios, increasing both the robustness and safety of \ac{APhI} tasks.
%
%Further developments of the \ac{AM} will cover more complex interaction scenarios, exploiting the high range of possible compliance values such to increase both robustness and safety of \ac{APhI} tasks.
%
We will also further address the problem of the critical folding configurations, leading to a more robust mechanical design.
%
%Also, the high range of possible compliance values pave the way to innovative control laws to further increase the robustness and safety of the physical interaction, 

%\textbf{Disadvantages of the approach to be further investigated? (still underactuated in roll, decentralized control, critical folding configurations)}.

% \addtolength{\textheight}{-11.2cm}   % This command serves to balance the column lengths

%%%%%%%%%%%%%%%%%%%%%%%%%%%%%%%%%%%%%%%%%%%%%%%%%%%%%%%%%%%%%%%%%%%%%%%%%%%%%%%%

%%%%%%%%%%%%%%%%%%%%%%%%%%%%%%%%%%%%%%%%%%%%%%%%%%%%%%%%%%%%%%%%%%%%%%%%%%%%%%%%

%%%%%%%%%%%%%%%%%%%%%%%%%%%%%%%%%%%%%%%%%%%%%%%%%%%%%%%%%%%%%%%%%%%%%%%%%%%%%%%%
\section*{ACKNOWLEDGMENT}
We thank Christoph Gaupp for his help on building the \ac{MAV} platform prototype.

%%%%%%%%%%%%%%%%%%%%%%%%%%%%%%%%%%%%%%%%%%%%%%%%%%%%%%%%%%%%%%%%%%%%%%%%%%%%%%%%

\bibliographystyle{IEEEtran}
\bibliography{miniVoliroBibs}

\end{document}

%% file: symbol_definitions.tex
\usepackage{acro}
\usepackage{bm}
\usepackage{mathtools}
\usepackage{amsfonts}
\usepackage{amsmath}
\usepackage{xcolor}
\usepackage{amssymb}
\usepackage[load-configurations = abbreviations]{siunitx}
\usepackage{caption}
\usepackage{subcaption}
\let\labelindent\relax
\usepackage[inline]{enumitem}

\DeclareAcronym{ASL}{short = ASL, long = Autonomous Systems Lab}
\DeclareAcronym{OMAV}{short = OMAV, long = Omnidirectional Micro Aerial Vehicle}
\DeclareAcronym{MAV}{short = MAV, long = Micro Aerial Vehicle}
\DeclareAcronym{DOF}{short = DOF, long = degrees of freedom}
\DeclareAcronym{PBC}{short = PBC, long = passivity-based control}
\DeclareAcronym{PH}{short = PH, long = Port-Hamiltonian}
\DeclareAcronym{NDT}{short = NDT, long = non-destructive testing}
\DeclareAcronym{PEMS}{short = PEMS, long = Power and Energy Monitoring System}
\DeclareAcronym{WTC}{short = WTC, long = wrench tracking controller}
\DeclareAcronym{MBE}{short = MBE, long = momentum-based wrench estimator}
\DeclareAcronym{ASIC}{short = ASIC, long = Axis-Selective Impedance Control}
\DeclareAcronym{MPC}{short = MPC, long = Model Predictive Control}
\DeclareAcronym{APhI}{short = APhI, long = Aerial Physical Interaction}
\DeclareAcronym{LLE}{short = LLE, long = Largest Lyapunov Exponent}
\DeclareAcronym{ICBF}{short = ICBF, long = Integral Control Barrier Functions}
\DeclareAcronym{CBF}{short = CBF, long = Control Barrier Functions}
\DeclareAcronym{COM}{short = CoM, long = Center of Mass}
\DeclareAcronym{AM}{short = AM, long = Aerial Manipulator}
\DeclareAcronym{IK}{short = IK, long = Inverse Kinematics}
\DeclareAcronym{FT}{short = F/T, long = force and torque, short-indefinite = an, long-indefinite = a}

\newcommand{\EC}[1]{#1}

\renewcommand{\vec}[1]{\bm{#1}}		% vectors
\newcommand{\matr}[1]{\bm{#1}}		% matrices
\newcommand{\mat}[1]{\bm{#1}}		% matrices
\newcommand{\nR}[1]{\mathbb{R}^{#1}}		% real number
\renewcommand{\matrix}[1]{\begin{bmatrix} #1 \end{bmatrix}}	% matrix
\newcommand{\upperRomannumeral}[1]{\uppercase\expandafter{\romannumeral#1}}	% roman numbers

\newcommand{\vSpace}{\;\,}
\newcommand{\transpose}{^\top}
\newcommand{\refer}{^{\text{ref}}}
\newcommand{\W}[1]{\prescript{}{W}{#1}}
\newcommand{\B}[1]{\prescript{}{B}{#1}}

\newcommand{\E}[1]{\prescript{}{E}{#1}}
\newcommand{\D}[1]{\prescript{}{D}{#1}}

\renewcommand{\frame}{\mathcal{F}}		% frame
\newcommand{\pos}{\vec{p}}				% position vector
\newcommand{\vel}{\vec{v}}				% velocity vector
\newcommand{\acc}{\dot{\vel}}				% acceleration vector

\newcommand{\rotMat}{\matr{R}}				% rotation matrix
\newcommand{\frameW}{\frame_W}			% world frame
\newcommand{\frameB}{\frame_B}			% body frame
\newcommand{\frameD}{\frame_D}			% L frame
\newcommand{\frameE}{\frame_E}			% L frame
\newcommand{\x}{\vec{x}}
\newcommand{\y}{\vec{y}}
\newcommand{\z}{\vec{z}}

\newcommand{\rotMatWB}{\rotMat_{WB}}	% rotation matrix for load
\newcommand{\rotMatBW}{\rotMat_{BW}}	% rotation matrix for load orientation
\newcommand{\rotMatDB}{\rotMat_{DB}}	% rotation matrix for load orientation
\newcommand{\angVel}{\vec{\omega}}
\newcommand{\angAcc}{\dot{\vec{\omega}}}

\DeclarePairedDelimiter{\norm}{\lVert}{\rVert} % norm
\DeclareMathOperator{\atanTwo}{atan2}
\DeclareMathOperator{\sign}{sign}

\newcommand{\inertia}{\bm{J}}

\newcommand{\mass}{m}

\newcommand{\Kstiffness}{k_s}

\newcommand{\error}{\bm{e}}

\newcommand{\positionError}{\error_p}

\newcommand{\velocityError}{\error_v}

\newcommand{\Kmat}{\mat{K}}

\newcommand{\Dmat}{\mat{D}}
\newcommand{\kpPos}{\Kmat_p}

\newcommand{\kdPos}{\Dmat_p}

\newcommand{\forceCommand}{\vec{f}_{c}}
\newcommand{\reducedForceCommand}{\vec{f}^r_{c}}
\newcommand{\forceCommandi}[1]{f_{c,#1}}
\newcommand{\underForceCommand}{\bar{\vec{f}}_{c}}
\newcommand{\torqueCommand}{\vec{\tau}_{c}}

\newcommand{\angError}{\error_{R}}

\newcommand{\angVelError}{\error_{\omega}}
\newcommand{\kpAng}{\Kmat_R}
\newcommand{\kdAng}{\Dmat_\omega}

\newcommand{\gravityForce}{\vec{f}_g}
\newcommand{\inputSin}{\vec{u}}

\newcommand{\lv}{\vec{l}}
\newcommand{\Pv}{\vec{p}}
\newcommand{\des}{^d}

\newcommand{\eefPose}{\pos_e}
\newcommand{\eefPoseRef}{\pos\refer_e}
\newcommand{\eefPoseDes}{\pos^d_e}
\newcommand{\deltaJoint}{\theta_i}